\documentclass[10pt,twocolumn,letterpaper,table]{article}

\usepackage{cvpr}
\usepackage{times}
\usepackage{epsfig}
\usepackage{graphicx}
\usepackage{amsmath}
\usepackage{amssymb}
\usepackage{subcaption}
\usepackage{threeparttable}
\usepackage{booktabs}
\usepackage{adjustbox}
\usepackage{xcolor}
\usepackage{pbox}
\usepackage{dsfont}
\usepackage{pifont}
\usepackage{enumerate}
\usepackage{subfloat}
\usepackage{mathabx}
\usepackage{pgfplots}
\usepackage{color}
\usepackage{multirow}
\usepackage{algorithm}
\usepackage{algorithmic}

\graphicspath{{images/}}

\definecolor{Gray}{gray}{0.9}

\usepackage[pagebackref=true,breaklinks=true,letterpaper=true,colorlinks,bookmarks=false]{hyperref}

\cvprfinalcopy 

\def\cvprPaperID{1201} 

\ifcvprfinal\fi
\begin{document}

\title{GSTO: Gated Scale-Transfer Operation for Multi-Scale Feature Learning in Pixel Labeling}

\author{
    Zhuoying Wang$^{1}$, Yongtao Wang$^{1}$\thanks{Corresponding author},~Zhi Tang$^{1}$,\\
    Yangyan Li$^{2}$, Ying Chen$^2$, Haibin Ling$^{3}$,Weisi Lin$^{4}$\\
    $^1$Peking University, 
    $^2$Alibaba Group\\
    $^3$Stony Brook University, $^4$Nanyang Technological University\\
    {\tt \small
    \{wzypku, wyt, tangzhi\}@pku.edu.cn, \{chenying.ailab, yangyan.lyy\}@alibaba-inc.com} \\
    {\tt \small hling@cs.stonybrook.edu, wslin@ntu.edu.sg}
}

\maketitle

\begin{abstract}
  Existing CNN-based methods for pixel labeling heavily depend on multi-scale features to meet the requirements of both semantic comprehension and detail preservation. State-of-the-art pixel labeling neural networks widely exploit conventional scale-transfer operations, i.e., up-sampling and down-sampling to learn multi-scale features. In this work, we find that these operations lead to scale-confused features and suboptimal performance because they are spatial-invariant and directly transit all feature information cross scales without spatial selection. To address this issue, we propose the Gated Scale-Transfer Operation (GSTO) to properly transit spatial-filtered features to another scale. Specifically, GSTO can work either with or without extra supervision. Unsupervised GSTO is learned from the feature itself while the supervised one is guided by the supervised probability matrix. Both forms of GSTO are lightweight and plug-and-play, which can be flexibly integrated into networks or modules for learning better multi-scale features. In particular, by plugging GSTO into HRNet, we get a more powerful backbone (namely GSTO-HRNet) for pixel labeling, and it achieves new state-of-the-art results on the COCO benchmark for human pose estimation and other benchmarks for semantic segmentation including 
  Cityscapes, LIP and Pascal Context, with negligible extra computational cost. Moreover, experiment results demonstrate that GSTO can also significantly boost the performance of multi-scale feature aggregation modules like PPM and ASPP. Code will be made available at https://github.com/VDIGPKU/GSTO.
\end{abstract}

\begin{figure*}
  \includegraphics[width=\textwidth]{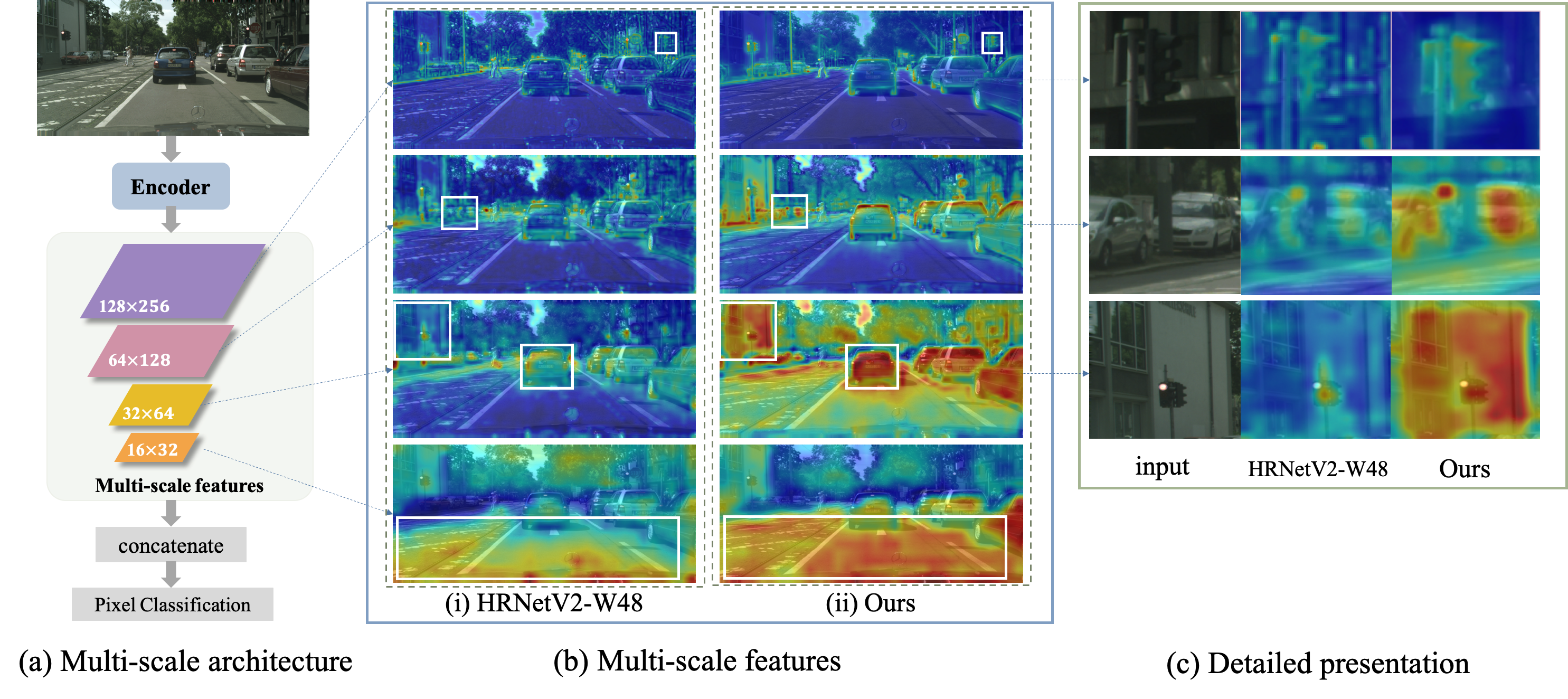}
  \caption{Visual comparison of the multi-scale features extracted by the encoder of (i) HRNetV2-W48 and (ii) our proposed GSTO-HRNet. Each heat map is obtained by averaging the corresponding feature map along the channel dimension, and warmer color (red) indicates larger activation. The comparison demonstrates that our approach obtains more discriminate and scale-aware features, where small objects like ``traffic light” and object boundaries are more precisely highlighted in the high-resolution feature map, while medium-size objects like “car” and far-away “building” as well as large objects like “road” and nearby “car” are better focused in low-resolution feature maps. On the contrast, HRNetV2 suffers from feature-confusion, that is, some parts of large objects incorrectly fire high activation responses on the high-resolution features and large objects are insufficiently focused on the low-resolution features.}
  \label{fig:introduction}
\end{figure*}
\section{Introduction}

Pixel labeling tasks, such as semantic segmentation and human pose estimation, target at assigning contextual labels for each pixel of an image, and are requested to deal with classification and localization simultaneously~\cite{peng2017large}. Since classification requires a large receptive field for inferring the semantic category while localization requires high-resolution details for outlining the precise boundary, how to meet both the requirements is essential for the design of dense-pixel labeling algorithms~\cite{chen2014semantic}. 

Current state-of-the-art pixel-labeling methods generally exploit multi-scale features to handle the aforementioned issue and have obtained impressive results. Ideally, multi-scale features work by assigning pixels to a proper receptive field according to its positions and object scales, but in practice the learned features are often scale-confused.
An example is shown in Figure~\ref{fig:introduction}(a), where the multi-scale features are extracted from an image of Cityscapes \texttt{val} dataset by HRNetV2-W48~\cite{sun2019high}, one of the most powerful backbones for semantic segmentation.
One can observe that, in general, on the high-resolution feature map with small receptive field, small objects (e.g., “person” and “traffic light”) and the boundaries of large objects are highlighted, while on the low-resolution feature map with large receptive field, larger objects like “car” and “road” are stressed. Such observations show that the learned multi-scale features are mainly scale-aware, that is, the high-resolution features are responsible for sensing small objects and boundaries, while the low-resolution features are concerned with large objects. However, if investigating more carefully, we can find that the features learned by HRNetV$2$-W$48$ are not sufficiently scale-ware, that is, some parts of large objects incorrectly fire high activation responses on the high-resolution features and large objects are insufficiently focused on the low-resolution features. 

In this work, for the first time, we show that such scale-confusion is attributed to the spatial-invariant scale-transfer operations (i.e., up-sampling and down-sampling) that are extensively exploited by existing pixel labeling methods when learning multi-scale features. These operations  directly transit all feature information cross scales without scale-aware selection, leading to suboptimal performance.

To alleviate the above scale-confusion and learn scale-aware features for pixel labeling, we propose novel Gated Scale-Transfer Operations (GSTO) of two forms, unsupervised GSTO and supervised GSTO, to properly transit a feature map across scale. Specifically, unsupervised GSTO directly produces a pixel-wise gating map from the feature map itself, while supervised GSTO learns the gating map with supervision during the training phase. The proposed two GSTOs are lightweight and plug-and-play, thus it can be flexibly integrated into networks or modules for learning better multi-scale features at only minor extra computation cost. By replacing conventional scale-transfer operation with the proposed GSTO in HRNet~\cite{sun2019deep,sun2019high}, our approach, denoted as GSTO-HRNet, enjoys much more discriminative features for each scale. As shown in Figure~\ref{fig:introduction}(b), object boundary is more precisely outlined on high-resolution feature maps, and large objects are better focused on low-resolution feature maps. Quantitatively, GSTO-HRNet achieves new state-of-the-art results for human pose estimation on the COCO dataset with only a half amount of parameters and FLOPs, and for semantic segmentation on the Cityscapes, LIP and Pascal Context datasets with negligible extra computational costs. Extensive results show that the proposed GSTO can also improve modules for multi-scale feature aggregation modules like Pyramid Pooling Module~\cite{zhao2017pyramid} and Atrous Spatial Pyramid Pooling Module~\cite{chen2018deeplab} by a large margin.

In summary, our contributions are three-fold: 
\begin{itemize}
\item We propose two novel light-weight Gated Scale-Transfer Operations (GSTOs), unsupervised GSTO and supervised GSTO, to learn better multi-scale features for pixel labeling.
\item By plugging GSTOs into HRNet, we further propose a backbone named GSTO-HRNet and achieve new state-of-the-art results on multiple benchmarks for both semantic segmentation and human pose estimation.
\item The proposed GSTOs can also significantly improve the performance of modules for multi-scale feature aggregation modules.
\end{itemize}

\section{Related Work}
\subsection{Pixel labeling networks}
Pixel labeling tasks like semantic image segmentation and human pose estimation, require the capturing of both high-level semantic category and low-level spatial details. Though current CNN-based methods~\cite{pohlen2017full,ChenZPSA18,LinMSR17} reduce down-sampling layers to keep high-resolution~\cite{pohlen2017full} and exploit dilated convolution~\cite{chen2017rethinking} as well as large-kernel~\cite{peng2017large} convolution to expand the receptive field, multi-scale feature exploiting is still the most effective way to handle the above problem. Multi-scale aggregation modules~\cite{zhao2017pyramid,chen2018deeplab,Yang_2018_CVPR} are introduced at the end of encoder to extract features of various receptive fields.  Multi-stage networks~\cite{newell2016stacked,Cheng_2019_ICCV} are further exploited to processively combine semantic information and spatial details. Recently, an efficient and powerful backbone HRNet~\cite{sun2019deep,sun2019high} is proposed to process multi-scale features in parallel, reaching the best results on multiple benchmarks.

\subsection{Scale-transfer Operations}
Conventional scale-transfer operations like average pooling and bilinear interpolation are widely used as cross-scale transition methods in current multi-scale feature aggregation modules~\cite{chen2017rethinking,zhao2017pyramid,Yang_2018_CVPR,TakikawaAJF19} and multi-scale feature extraction backbones~\cite{sun2019deep,sun2019high} for pixel labeling tasks. Besides, a few other transfer operations have been proposed. For example, \cite{shi2016real} proposes an efficient sub-pixel convolution layer to learn an array of upscaling filters and upscale the low-resolution features. \cite{tian2019decoders} introduces a data-dependent up-sampling method to replace the bilinear in decoders for semantic segmentation. 
Generally speaking, all of the aforementioned methods are designed only for upscaling, not plug-and-play, and suffer from heavy computational costs.    

\subsection{Gate Mechanism}
Gate mechanism has been widely exploited in computer vision to enhance the representational power by modeling channel-wise or spatial-wise relationship.~\cite{WangJQYLZWT17,ParkWLK18,hu2018squeeze,woo2018cbam} In pixel labeling tasks, self-attention mechanism is proposed to use the weighted combination of pixles or channels as the context. \cite{chen2016attention} designs a network to learn gates to ensemble multi-scale results at the end of model. Gated-SCNN\cite{TakikawaAJF19} proposed a two-stream CNN architecture and utilize gate mechanism to wire shape information as a seperate processing branch. Inspired by the works above but different, we argue that \textit{inserting heavy attention modules after each block or the whole backbone brings limited improvement}, and we further propose a light-weight gate mechanism and equip it with scale-transfer operations, significantly improving the performance of multiple multi-scale feature extraction methods.

\section{Method}
In this section, we first introduce the principle of our proposed Gated Scale-Transfer Operation (GSTO) and its two forms in Section~\ref{sec3.1}. Then we illustrate how to equip multi-scale backbones with GSTO and describe the pipeline of the advanced backbone GSTO-HRNet in Section~\ref{sec3.2}. Lastly in Section~\ref{sec3.3}, we show how to improve general multi-scale feature aggregation modules by utilizing GSTO with an example.

\subsection{Gated Scale-Transfer Operation}
\label{sec3.1}
\begin{figure}[t]
\small
\centering
\includegraphics[width=1.0\linewidth]{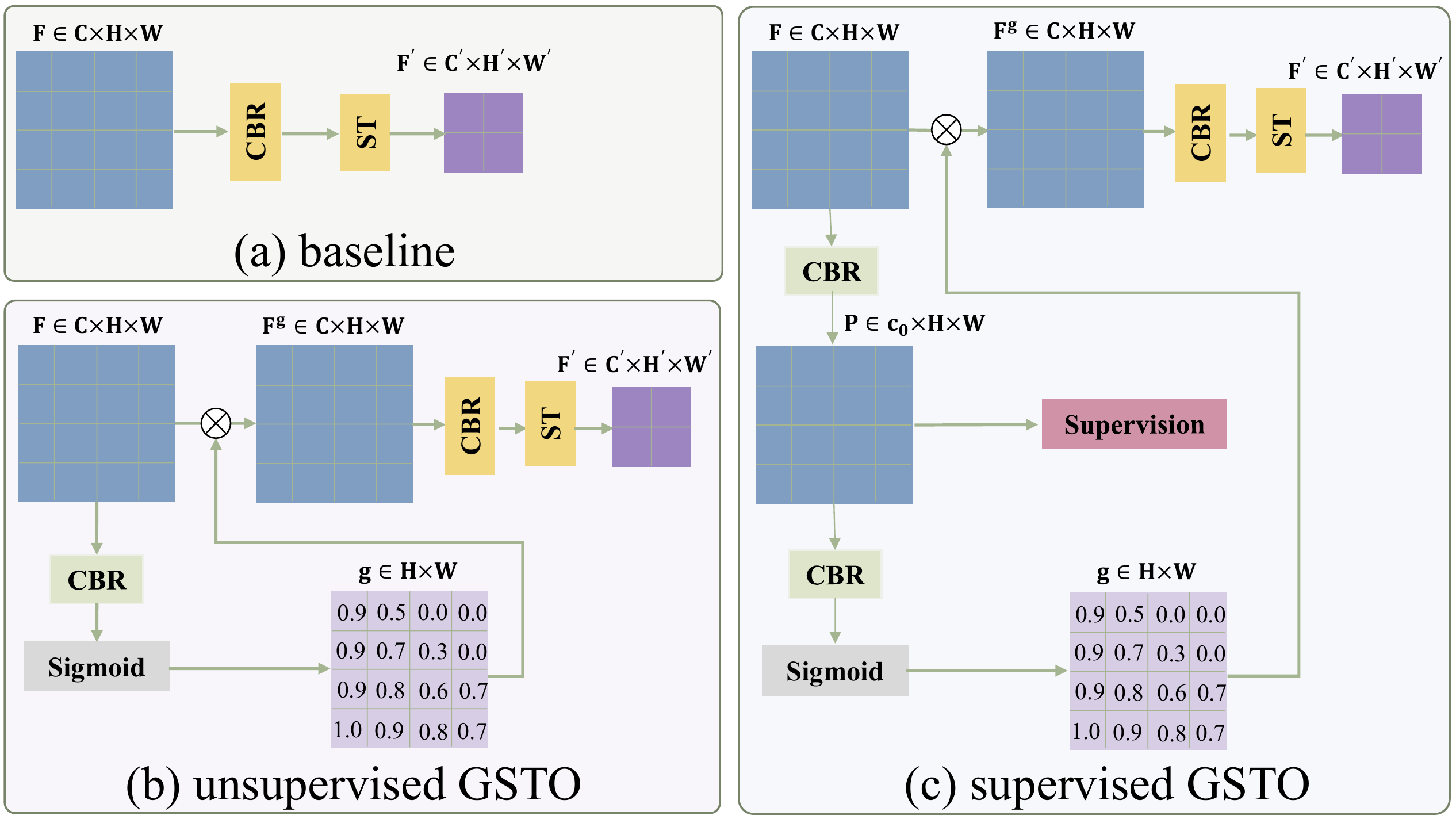}\\
\vspace{-.2cm}
\caption{\small Structures of the proposed unsupervised GSTO and supervised GSTO. \textbf{CBR} represents Conv+BN+ReLU, used to change the channel size, if needed, and \textbf{ST} refers to conventional scale-transfer operation including down-sampling and up-sampling.
}
\label{fig:GSTO}
\vspace{-.3cm}
\end{figure}
\begin{figure*}[t]
\footnotesize
    \centering
    \includegraphics[width = 0.99\textwidth]{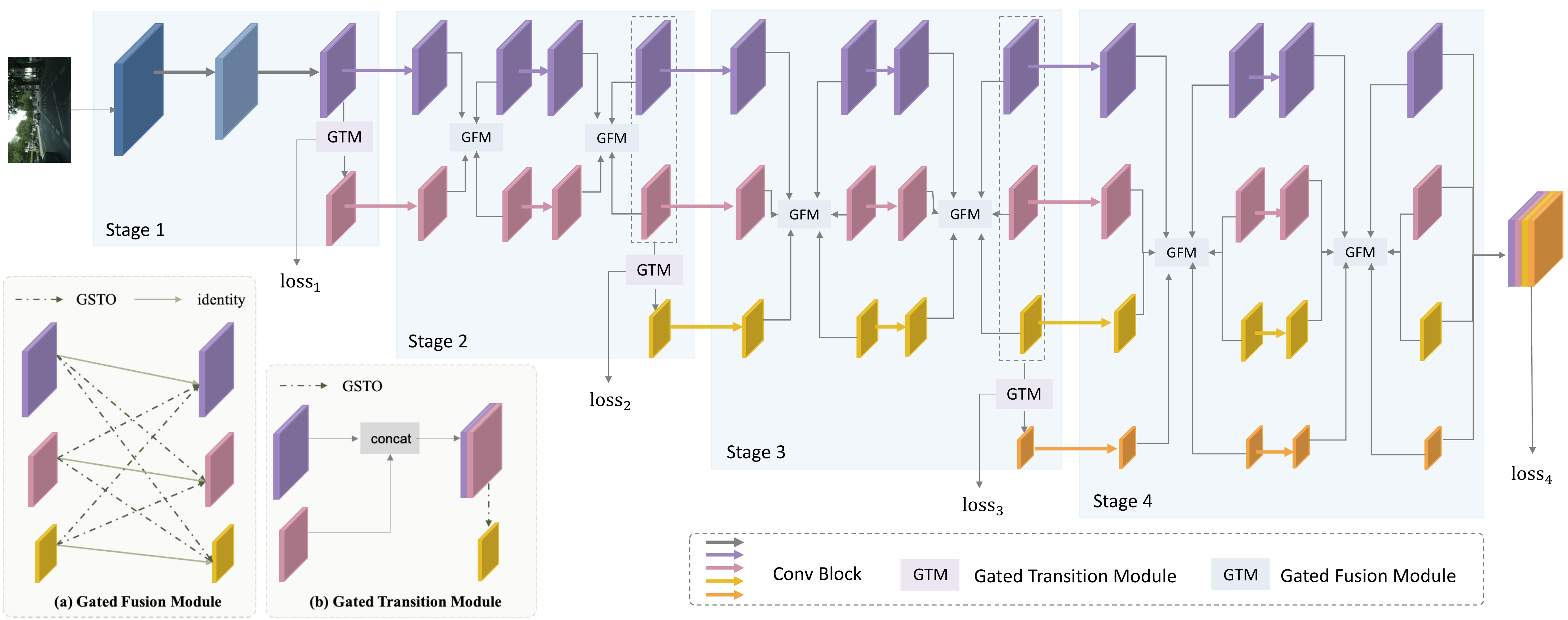}
    \caption{The pipeline of GSTO-HRNet, the GSTO-advanced multi-scale backbone(described in Section~\ref{sec3.2}). The GFM and GTM are GSTO-based modules for multi-scale feature fusion and generation, respectively.}
    \label{fig:pipeline}
    \vspace{-2mm}
\end{figure*}
\begin{figure}[t]
\small
\centering
\includegraphics[width=0.9\linewidth]{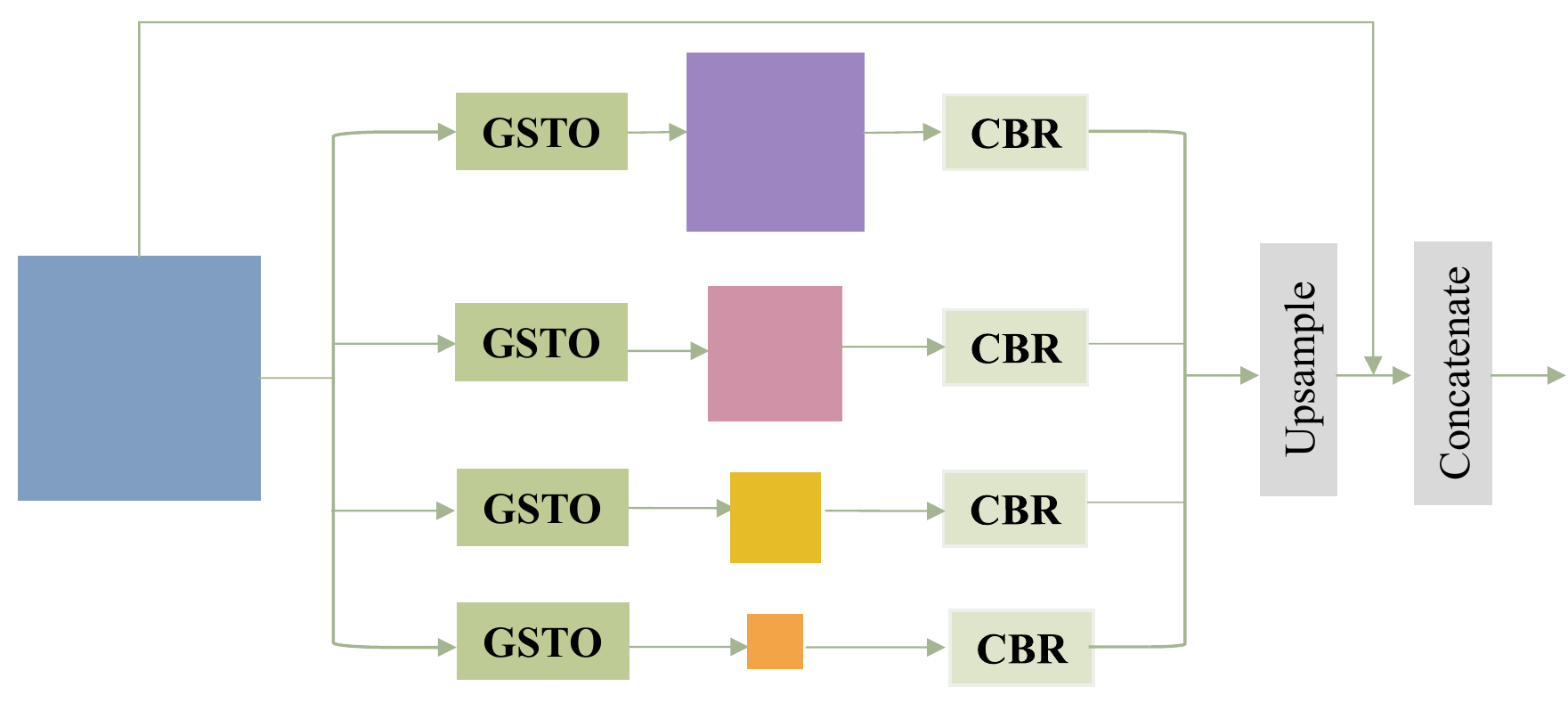}\\
\vspace{.1cm}
\caption{\small GSTO-based Pyramid Pooling Module: an example to advance multi-scale aggragation modules with the proposed GSTO. \textbf{CBR} represents Conv+BN+ReLU.
}
\label{fig:psp}
\vspace{-.5cm}
\end{figure}
The intuition of Gated Scale-Transfer Operation is to learn a spatial mask that filters pixels inconsistent with the target scale. The operation works during the cross-scale feature transition. We denote the initial feature to be transited as $F \in \mathbb{R}^{C \times H \times W}$, with $C$ channels of size $(W,H)$, and the target feature as $F{'} \in \mathbb{R}^{C{'} \times H{'} \times W{'}}$, with $C'$ channels of size $(W',H')$. The feature vector at location $(i,j)$ ($i=1,...,H$, $j=1,...,W$) is denoted as $F_{ij}\in
\mathbb{R}^{C}$, and similar notation is used for $F'$.

\subsubsection{Traditional Scale-Transfer Operation}

As shown in Figure~\ref{fig:GSTO}(a), traditional transition is performed through down-sampling like average pooling and up-sampling like bilinear interpolation. Additionally, if $C \ne C'$, convolutional layers are needed for channel modification. The process can be represented as follows:

\begin{equation}
\label{equ:1}
\widetilde{F}_{kij} = \sum_{m=1}^{C}\omega_{km}\cdot F_{mij},~~~~~k=1,...,C{'},
\end{equation}

\begin{equation}
\label{equ:2}
F{'} = ST(\widetilde{F}),
\end{equation}
where $\widetilde{F} \in \mathbb{R}^{C{'}\times H\times W}$ is computed by a $1\times 1$ convolution,  $\omega_{k}\in \mathbb{R}^{C}$ is the $k$-$th$ convolutional kernel, and $ST$ represents traditional scale-transfer operations.

 \subsubsection{Gated Scale-Transfer Operation}

In the proposed GSTOs (Figure~\ref{fig:GSTO}(b) and (c)), a spatially gated feature $F^{g}$ is produced firstly and then Equations~\ref{equ:1} and~\ref{equ:2} are performed on $F^{g}$ instead of on the original $F$. The element of $F^{g}$ is calculated by element-wise multiplication as follows:

 \begin{equation}
 \label{equ:3}
  F_{mij}^{g}=g_{ij}\cdot F_{mij},~~~~~m=1,...,C,
 \end{equation}
where $g_{ij}\in \mathbb{R}$ is the corresponding value of the gate at location $(i,j)$. 

For \textbf{unsupervised GSTO} (see Figure~\ref{fig:GSTO}(b)), element of the gate $g_{ij}$ is calculated from the original feature $F$, by an $1\times 1$ convolution with input channel of $C$ and output channel of $1$, followed by sigmoid, which can be denoted as:
\begin{equation}
\label{equ:4}
g_{ij}=\sigma(\sum_{m=1}^{C}\rho _{m} \cdot F_{mij}),
\end{equation}
where $\rho\in \mathbb{R}^{C}$ is the weight of the convolution, and $\sigma(\cdot)$ is the \textit{sigmoid} function defined as $\sigma(x)=(1+e^{-x})^{-1}$.

As for \textbf{supervised GSTO} (see Figure~\ref{fig:GSTO}(c)), a light-weight predictor, such as a $1\times 1$ convolution, is performed on $F$ to get $P\in \mathbb{R}^{c_{0}\times H\times W}$, where $c_{0}$ is the number of semantic categories and $P$ is supervised by the ground truth during training process. $P_{nij}$ measures the probability that pixel $(i,j)$ belongs to the $n$-$th$ class. Then we apply a $1\times 1$ convolution on $P$ to get the spatial mask. The process is represented mathematically as follows:

\begin{equation}
P_{nij}= \sum_{m=1}^{C}\omega_{nm}'\cdot F_{mij},~~~~~n=1,...,c_0,
\end{equation}

\begin{equation}
g_{ij}=\sigma(\sum_{n=1}^{c_0}\theta _{n} \cdot P_{nij}),
\end{equation}
that is to say, each element of the learned $\theta\in \mathbb{R}^{c_{0}}$ corresponds to a semantic category and represents the weight of this category when transferred to the target scale. 

\subsection{Multi-scale Backbone with GSTO}
\label{sec3.2}

The recently proposed multi-scale backbone HRNet~\cite{sun2019high,sun2019deep} has shown impressive results in pixel labeling. With our proposed GSTO, we build an advanced backbone named GSTO-HRNet (the pipeline is shown in Figure~\ref{fig:pipeline}). We focus on the multi-scale feature fusion after each block and the lower-resolution branch generation after each stage, and introduce corresponding GSTO-based modules: Gated Fusion Module (GFM) and Gated Transition Module (GTM).

\subsubsection{Gated Fusion Module}
In the design of GFM (Figure~\ref{fig:pipeline}(a)), we follow the densely-connected pattern in HRNet~\cite{sun2019deep} but replace the traditional cross-scale interaction with the proposed Gated Scale-Transfer Operation. Since GFM is performed after every block, we exploit unsupervised GSTO for efficiency. 

\subsubsection{Gated Transition Module}
Lower-resolution represents larger receptive field. HRNet cuts down the feature resolution by half after each stage through a $3\times 3$ stride convolution layer performed on the current lowest-resolution feature map. But for dense-pixel tasks, we prefer to keep the pixels that have been semantically comprehended in the high-resolution branch while transit those requiring larger receptive filed to the lower resolution one. So we adopt GSTO to achieve the selection. In the proposed Gated Transition Module~(Figure~\ref{fig:pipeline}(b)), we up-sample the features from each branch to the same resolution and concatenate them along the channel dimension. Then GSTO is performed on the united feature to get a lower-resolution branch. We will experimentally compare the unsupervised GSTO and supervised GSTO in Section~\ref{sec:4.3}. When we use supervised GSTO, the final loss is set as $0.2\times loss_1+0.3\times loss_2+0.5\times loss_3+1.0\times loss_4$, where $loss_i$ is the cross-entropy loss of the $i$-$th$ stage.

\subsection{Multi-scale Modules with GSTO}
\label{sec3.3}
Besides the above mentioned multi-scale backbones, traditional classification backbone can gain improvement by applying GSTO to multi-scale aggregation modules like Pyramid Pooling Module (PPM)~\cite{zhao2017pyramid} and Atrous Spatial Pyramid Pooling (ASPP)~\cite{chen2018deeplab}. Figure~\ref{fig:psp} shows an example of PPM advanced by GSTO. Generally, GSTO can be adopted to replace the conventional scale-transfer operations as adaptive pooling or atrous convolution to expand the receptive field.

\section{Experiment}
\setlength{\tabcolsep}{6.0pt}
	\begin{table}[t]
		\scriptsize
		\centering
		\begin{tabular}{l|c}
			\hline\noalign{\smallskip}
			  Method& mIoU\\
			 \hline
			

			\hline
			baseline~\cite{sun2019high} & $75.9$(\textit{impl}) \\
			baseline(w/ sup) & $76.3(0.4\uparrow)$ \\
			baseline+GFM & $76.6(0.7\uparrow)$ \\
			baseline+GTM(w/o sup) & $76.4(0.5\uparrow)$ \\
			baseline+GTM(w/ sup) & $77.0(1.1\uparrow)$ \\
			baseline+GTM(w/o sup)+GFM & $76.8(0.9\uparrow)$ \\
			baseline+GTM(w/ sup)+GFM & $\textbf{77.2}(1.3\uparrow)$ \\
			\hline
		\end{tabular}
		\caption{Comparison experiments of GFM and GTM.}
		\label{tab:ablation_hrnet1}
	\end{table}
\setlength{\tabcolsep}{6.0pt}
	\begin{table}[t]
		\scriptsize
		\centering
		\begin{tabular}{rrr|rr|c}
			\hline\noalign{\smallskip}
			 Stage1& Stage2& Stage3& \#Param. & GFLOPs & mIoU\\
			 \hline
			
			 \hline
			\ding{55}& \ding{55}& \ding{55}& $3.93$M & $71.7$ & $76.6$ \\
			\ding{55}& \ding{55}& \ding{51} & $3.95$M & $73.9$ & $76.7$ \\
			\ding{55}& \ding{51}& \ding{51} & $3.95$M & $74.4$ & $\textbf{77.3}$ \\
			\ding{51}& \ding{51}& \ding{51} & $4.02$M & $83.0$ & $77.2$ \\
			\hline
		\end{tabular}
		\caption{Ablation study for the number of GTM. The base model is HRNetV$2$ with GFM only(the first row).}
		\label{tab:ablation_hrnet2}
	\end{table}
\setlength{\tabcolsep}{6.0pt}
	\begin{table}[t]
		\scriptsize
		\centering
		\begin{tabular}{l|l|lc|c}
			\hline\noalign{\smallskip}
			Attention form& Location & \#Param. & GFLOPs & mIoU\\
			 \hline
			
			 \hline
			channel-wise& after each block  & $3.93$M & $71.6$ & $75.9$ \\
			spatial-wise& after each block  & $3.92$M & $71.7$ & $76.2$ \\
			channel-wise& after each layer  & $3.94$M & $71.6$ & $76.1$ \\
			spatial-wise& after each layer  & $3.93$M & $72.1$ & $76.4$ \\
			\hline
			channle-wise& combined with ST  & $3.94$ & $71.6$ & $76.2$ \\
			spatial-wise& combined with ST  & $3.93$M & $71.7$ & $\textbf{76.8}$ \\
			
			\hline
		\end{tabular}
		\caption{Comparison of different gating/attention mechanism. ST means Scale-transfer operation. All the experiments are conducted without extra supervision.}
		\label{tab:ablation_hrnet5}
	\end{table}
	\renewcommand{\arraystretch}{1.3}
	\setlength{\tabcolsep}{3pt}
	\begin{table}[t]
 		\scriptsize
		\centering
		\begin{tabular}{l|c|lc|cc|c}
			\hline\noalign{\smallskip}
			 Method &Backbone& \#Param.& incre. & GFLOPs & incre.& mIoU\\
			\hline

			\hline
			HRNetV$2$& HRNetV$2$-W$18$& $3.92$M& \multirow{2}{*}{$\blacktriangle 0.67\%$}& $71.6$& \multirow{2}{*}{$\blacktriangle 3.9\%$}& $76.2/75.9$(\textit{impl.})\\
			Ours& GSTO-HRNet-W$18$&$3.95$M& & $74.4$& &$\textbf{77.3}$($1.1/1.4\uparrow$)\\
			\hline
			HRNetV$2$& HRNetV$2$-W$48$& $65.78$M& \multirow{2}{*}{$\blacktriangle 0.23\%$}& $696.2$& \multirow{2}{*}{$\blacktriangle 2.6\%$}& $80.9/80.2$(\textit{impl.})\\
			Ours& GSTO-HRNet-W$48$&$65.93$M& & $714.0$& &$\textbf{82.1}$($1.2/1.9\uparrow$)\\
			\hline

			\hline
		\end{tabular}
		\caption{The increments of parameters and GFLOPs from HRNetV$2$ to our GSTO-HRNet and the mIoU comparison on Cityscapes \texttt{val.}(single scale and no flipping, not using OHEM during training).}
		\label{tab:ablation_hrnet3}
	\end{table}
	\renewcommand{\arraystretch}{1.3}
	\setlength{\tabcolsep}{3pt}
	\begin{table}[t]
        \scriptsize
		\centering
		\begin{tabular}{l|ccc|l}
			\hline\noalign{\smallskip}
			  &use val.& OHEM & MS  & mIoU\\
			\hline

			\hline
			HRNetV$2$-W$48$& \multirow{2}{*}{\ding{55}}& \multirow{2}{*}{\ding{55}}& \multirow{2}{*}{\ding{51}}& 
			$80.4$\\
			Ours-W$48$& & & & $\textbf{81.9}(1.5\uparrow)$\\
			\hline
			HRNetV$2$-W$48$& \multirow{2}{*}{\ding{51}}& \multirow{2}{*}{\ding{55}}& \multirow{2}{*}{\ding{51}}& 
			$81.5$\\
			Ours-W$48$& & & & $\textbf{82.3}(0.8\uparrow)$\\
			\hline
			HRNetV$2$-W$48$& \multirow{2}{*}{\ding{51}}& \multirow{2}{*}{\ding{51}}& \multirow{2}{*}{\ding{51}}& 
			$81.6$\\
			Ours-W$48$& & & & $\textbf{82.4}(0.8\uparrow)$\\

			\hline

			\hline
		\end{tabular}
		\caption{Comparison between HRNetV$2$ and GSTO-HRNet on Cityscapes \texttt{test}. \textit{use val.} means using \texttt{val.} set for training and \textit{MS} means Multi-scale testing.}
		\label{tab:ablation_hrnet4}
	\end{table}
	\renewcommand{\arraystretch}{1.3}
	\setlength{\tabcolsep}{3pt}
	\begin{table}[t]
		\scriptsize
		\centering
		\begin{tabular}{l|l|l}
			\hline\noalign{\smallskip}
			  Method& PPM& ASPP\\
			\hline

			\hline
			Baseline& $76.5$& $74.9$\\
			Baseline(w/ sup)& $76.9(0.4\uparrow)$& $75.1{(0.2\uparrow})$\\
			Baseline+GSTO(w/o sup)& $77.3(0.8\uparrow)$& $76.3(1.4\uparrow)$\\
			Baseline+GSTO(w/ sup)& $77.8(1.3\uparrow)$& $76.9(2.0\uparrow$)\\
			\hline

			\hline
		\end{tabular}
		\caption{Improvement on multi-scale aggregation modules.}
		\label{tab:ablation_ppm}
	\end{table}

\setlength{\tabcolsep}{6.0pt}
	\begin{table}[t]
		\scriptsize
		\centering
		\begin{tabular}{l|lcc|c}
			\hline\noalign{\smallskip}
			 & Backbone & \#Param. & GFLOPs & mIoU\\
			 \hline
			
			 \hline
			UNet++~\cite{zhou2018unet++} & ResNet-$101$ & $59.5$M & $748.5$ & $75.5$ \\
			DeepLabv3~\cite{chen2017rethinking} & Dilated-ResNet-$101$ & $58.0$M & $1778.7$ & $78.5$ \\
			DeepLabv3+~\cite{chen2018encoder} & Dilated-Xception-$71$ & $43.5$M & $1444.6$ & $79.6$ \\
			PSPNet~\cite{zhao2017pyramid} & Dilated-ResNet-$101$ & $65.9$M & $2017.6$ & $79.7$ \\
			ACFNet\cite{zhang2019acfnet} & ResNet-$101$ & - & - & $80.1$ \\
			SPGNet\cite{Cheng_2019_ICCV} & $2\times$ResNet-$50$ & $59.8$M & $654.8$ & $80.9$ \\
			HRNetV$2$\cite{sun2019high} & HRNetV$2$-W$48$ & $65.8$M & $696.2$ & $80.9$ \\
			\hline
			Our approach & GSTO-HRNet-W$48$ & $65.9$M & $714.0$ & $\textbf{82.1}$ \\
			\hline
		\end{tabular}
		\caption{Comparison of semantic segmentation results on
		Cityscapes \texttt{val.}
	    (single scale and no flipping, not using OHEM during training).
		The GFLOPs is calculated on the input size $1024 \times 2048$.}
		\label{tab:cityscapes_sota_val}
	\end{table}
\renewcommand{\arraystretch}{1.3}
	\setlength{\tabcolsep}{3pt}
	\begin{table}[t]
		\scriptsize
		\centering
		\begin{tabular}{l|l|cccc}
			\hline\noalign{\smallskip}
			 Method & Backbone & mIoU  & iIoU cla. & IoU cat. & iIoU cat.\\
			\hline
			
			\hline
			\multicolumn{3}{l}{\emph {Model learned on the \texttt{train} set}}\\
			\hline
			PSPNet~\cite{zhao2017pyramid} & Dilated-ResNet-$101$ & $78.4$ & $56.7$ & $90.6$  & $78.6$ \\
			PSANet~\cite{psanet} & Dilated-ResNet-$101$ & $78.6$ & - & - & - \\
			PAN~\cite{li2018pyramid} & Dilated-ResNet-$101$ & $78.6$ & - & - & - \\
			AAF~\cite{aaf2018} & Dilated-ResNet-$101$ & $79.1$ & - & - & -\\
			HRNetV$2$\cite{sun2019high} & HRNetV$2$-W$48$ & $80.4$ & $59.2$ & $91.5$ & $80.8$\\
			ACFNet\cite{zhang2019acfnet} &  ResNet-$101$  & $80.8$ & - & - & - \\
			\hline
			Our approach & GSTO-HRNet-W$48$ & $\mathbf{81.8}$ & $\mathbf{62.3}$ & $\mathbf{92.1}$ & $\mathbf{81.7}$\\
			\hline
			
			\hline
			\multicolumn{3}{l}{
			\emph {Model learned on the \texttt{train+valid} set}}\\
			\hline
			GridNet~\cite{fourure2017residual} & - & $69.5$ & $44.1$ & $87.9$ & $71.1$\\
			DeepLab~\cite{chen2017rethinking} & Dilated-ResNet-$101$ & $70.4$ & $42.6$ & $86.4$ & $67.7$\\
			FRRN~\cite{pohlen2017full}& - & $71.8$ & $45.5$ & $88.9$ & $75.1$\\
			RefineNet~\cite{lin2017refinenet}& ResNet-$101$ & $73.6$ & $47.2$ & $87.9$ & $70.6$\\
			DepthSeg~\cite{kong2018recurrent} & Dilated-ResNet-$101$ & $78.2$& - & - & - \\
			BiSeNet~\cite{yu2018bisenet} & ResNet-$101$ & $78.9$ & - & - & - \\
			DFN~\cite{yu2018learning} & ResNet-$101$ & $79.3$ & - & - & - \\
			PSANet~\cite{psanet} & Dilated-ResNet-$101$ & $80.1$ & - & - & - \\
			DenseASPP~\cite{Yang_2018_CVPR} & WDenseNet-$161$ & $80.6$ & $59.1$ & $90.9$ & $78.1$ \\
			SPGNet\cite{Cheng_2019_ICCV} &  $2\times$ResNet-$50$ & $81.1$ & - & - & - \\
			HRNetV$2$\cite{sun2019high} &  HRNetV$2$-W$48$  & $81.6$ & $61.8$ & $92.1$ & $82.2$ \\
			ACFNet\cite{zhang2019acfnet} &  ResNet-$101$  & $81.8$ & - & - & - \\
			\hline
			Our approach &  GSTO-HRNet-W$48$  & $\mathbf{82.4}$ & $\mathbf{63.8}$ & $\mathbf{92.4}$ & $\mathbf{83.3}$ \\
			\hline
		\end{tabular}
		\caption{Comparison with state-of-the-art segmentation results on Cityscapes \texttt{test}. }
		\label{tab:cityscaperesults}
	\end{table}
\renewcommand{\arraystretch}{1.3}
	\setlength{\tabcolsep}{2.8pt}
	\begin{table}[t]
	\scriptsize
	\centering
	\begin{tabular}{l|lc|ccc}
		\hline
		\noalign{\smallskip}
		 Method& Backbone & Extra. & Pixel acc. & Avg. acc. & mIoU \\
		\hline
		
		\hline
		Attention+SSL~\cite{gong2017look} & VGG$16$ & Pose & $84.36$ & $54.94$ & $44.73$ \\
		DeepLabV$3$+~\cite{chen2018encoder} & Dilated-ResNet-$101$ & - & $84.09$ & $55.62$ & $44.80$ \\
		MMAN~\cite{luo2018macro} & Dilated-ResNet-$101$ & - & - & - & $46.81$ \\
		SS-NAN~\cite{zhao2017self} & ResNet-$101$ & Pose & $87.59$ & $56.03$ & $47.92$ \\
		MuLA~\cite{nie2018mutual} & Hourglass & Pose & $88.50$ & $60.50$ & $49.30$ \\
		JPPNet~\cite{liang2018look} & Dilated-ResNet-$101$ & Pose & $86.39$ & $62.32$ & $51.37$ \\
		CE2P~\cite{ruan2019devil}  & Dilated-ResNet-$101$ & Edge & $87.37$ & $63.20$ & $53.10$ \\
		HRNetV$2$\cite{sun2019high} & HRNetV$2$-W$48$ & N & $88.21$ & $67.43$ & $55.90$ \\
		\hline
		Our approach & GSTO-HRNet-W$48$ & N & $\mathbf{88.38}$ & $\mathbf{68.36}$ & $\mathbf{57.37}$ \\	
		\hline
	\end{tabular}
	 \caption{Semantic segmentation results on LIP. $N$ denotes not using any extra information, e.g., pose or edge. }
	 \label{tab:lipresults}
	\end{table}
\renewcommand{\arraystretch}{1.3}
\begin{table*}[t]
\scriptsize
\centering
\begin{tabular}{l|l|c|r|c|cccccc|cccccc}
\hline
\multirow{2}{*}{Method} & \multirow{2}{*}{Backbone}  & \multirow{2}{*}{Input size} & \multirow{2}{*}{\#Param} & \multirow{2}{*}{GFLOPs} & \multicolumn{6}{c|}{Val} &\multicolumn{6}{c}{Test}\tabularnewline
\cline{6-17}
& & & & & $\operatorname{AP}$ & $\operatorname{AP}^{50}$ & $\operatorname{AP}^{75}$ & $\operatorname{AP}^{M}$ & $\operatorname{AP}^{L}$ & $\operatorname{AR}$ &$\operatorname{AP}$ & $\operatorname{AP}^{50}$ & $\operatorname{AP}^{75}$ & $\operatorname{AP}^{M}$ & $\operatorname{AP}^{L}$ & $\operatorname{AR}$\\
\hline
	\multicolumn{17}{c}{Bottom-up: keypoint detection and grouping}\\
\hline
OpenPose~\cite{CaoSWS17} & $-$& $-$ &$-$& $-$
				&$-$&$-$ &$-$ &$-$ &$-$ &$-$
				&$61.8$ & $84.9$&$67.5$&$57.1$&$68.2$&$ 66.5$\\
Associate Embedding~\cite{NewellHD17} & $-$& $-$ &$-$& $-$
				&$-$&$-$ &$-$ &$-$ &$-$ &$-$
				&$65.5$ & $86.8$&$72.3$&$60.6$&$72.6$&$ 70.2$\\
PersonLab~\cite{PapandreouZCGTM18} & $-$& $-$ &$-$& $-$
				&$-$&$-$ &$-$ &$-$ &$-$ &$-$
				&$68.7$ & $89.0$&$75.4$&$64.1$&$75.5$&$ 75.4$\\
MultiPoseNet~\cite{KocabasKA18} & $-$& $-$ &$-$& $-$
				&$-$&$-$ &$-$ &$-$ &$-$ &$-$
				&$69.6$ & $86.3$&$76.6$&$65.0$&$76.3$&$ 73.5$\\

\hline 
	\multicolumn{17}{c}{Top-down: human detection and single-person keypoint detection}\\
\hline
Mask-RCNN~\cite{HeGDG17} & ResNet-50-FPN& $-$ &$-$& $-$
				&$-$&$-$ &$-$ &$-$ &$-$ &$-$
				&$63.1$ & $87.3$&$68.7$&$57.8$&$71.4$&$ -$\\
G-RMI~\cite{PapandreouZKTTB17} & ResNet-101& $353\times257$ &$-$& $-$
				&$-$&$-$ &$-$ &$-$ &$-$ &$-$
				&$64.9$ & $85.5$&$71.3$&$62.3$&$70.0$&$ 69.7$\\
Integeral Pose Regression~\cite{SunXWLW18} & ResNet-101& $256\times256$ &$45.0$M& $11.0$
				&$-$&$-$ &$-$ &$-$ &$-$ &$-$
				&$67.8$ & $88.2$&$74.8$&$63.9$&$74.0$&$ -$\\
$8$-stage Hourglass~\cite{newell2016stacked} & $8$-stage Hourglass &  $256 \times 192$ & $25.1$M & $14.3$&
				$66.9$&$-$&$-$&$-$&$-$&$-$& 
				$-$&$-$ &$-$ &$-$ &$-$ &$-$ \\ 
CPN~\cite{chen2018cascaded}& ResNet-50 & $256 \times 192$ & $27.0$M & $6.20$
				&$68.6$&$-$&$-$&$-$&$-$&$-$
				&$-$&$-$ &$-$ &$-$ &$-$ &$-$\\ 
RMPE~\cite{FangXTL17} & PyraNet& $320\times256$ &$28.1$M& $26.7$
				&$-$&$-$ &$-$ &$-$ &$-$ &$-$
				&$72.3$ & $89.2$&$79.1$&$68.0$&$77.2$&$ 78.5$\\
CFN~\cite{HuangGT17} & ResNet-Inception& $384\times288$ &$-$& $-$
				&$-$&$-$ &$-$ &$-$ &$-$ &$-$
				&$72.6$ & $86.1$&$69.7$&$78.3$&$64.1$&$ -$\\
CPN (ensemble)~\cite{chen2018cascaded} & ResNet-Inception& $384\times288$ &$-$& $-$
				&$-$&$-$ &$-$ &$-$ &$-$ &$-$
				&$73.0$ & $91.7$&$80.9$&$69.5$&$78.1$&$ 79.0$\\
SimpleBaseline~\cite{xiao2018simple} & ResNet-152  & $256\times192$ &$68.6$M &$15.7$
&${72.0}$ & ${89.3}$&${79.8}$&${68.7}$&${78.9}$&${77.8}$\\
SimpleBaseline~\cite{xiao2018simple} & ResNet-152  &  $384\times288$     &$68.6$M &$35.6$
				&${74.3}$ & ${89.6}$&${81.1}$&${70.5}$&${79.7}$&${79.7}$
				&${73.7}$ & ${91.9}$&${81.1}$&${70.3}$&${80.0}$&${79.0}$\\
\hline
HRNet-W$32$\cite{sun2019deep} &HRNet-W$32$ &  $384\times 288$&  $28.5$M &$16.0$ & 
				$75.8$&$90.6$&${82.7}$&$71.9$&$82.8$&$81.0$  
				&$74.9$&$\textbf{92.5}$&$82.8$&$71.3$&$80.9$&$80.1$\\
HRNet-W$48$\cite{sun2019deep} & HRNet-W$48$&  $384\times 288$&  $63.6$M &$32.9$ & 
				$76.3$&$\textbf{90.8}$&$82.9$&$72.3$&$83.4$&$81.2$  
				& $75.5$&$\textbf{92.5}$&$83.3$&$71.9$&$81.5$&$80.5$\\
\hline
Our approach &GSTO-HRNet-W$32$ &  $384\times 288$&  \underline{$29.6$M} &\underline{$18.2$} & 
				$\textbf{76.5}$&$90.6$&$\textbf{83.1}$&$\textbf{72.6}$&$\textbf{83.7}$&$\textbf{81.4}$  
				&$75.5$&$92.4$&$83.2$&$\textbf{72.0}$&$81.5$&$\textbf{80.6}$\\
Our approach &GSTO-HRNet-W$48$ &  $384\times 288$&  $66.0$M &$37.6$ & 
				$\textbf{76.7}$&$90.7$&$\textbf{83.0}$&$\textbf{72.8}$&$\textbf{83.8}$&$\textbf{81.6}$  
				&$\textbf{75.8}$&$\textbf{92.5}$&$\textbf{83.4}$&$\textbf{72.3}$&$\textbf{81.8}$&$\textbf{80.9}$\\
\hline
\end{tabular}
\caption[Caption for LOF]{Comparisons on the COCO validation and test sets for pose estimation.}
\label{table:coco}
\end{table*}
\renewcommand{\arraystretch}{1.3}
	\setlength{\tabcolsep}{3.0pt}
	\begin{table}[t]
	\scriptsize
	\centering
	\begin{tabular}{l|l|cc}
		\hline\noalign{\smallskip}
		  Method& Backbone & mIoU ($59$ classes) & mIoU ($60$ classes) \\
		\hline
		
		\hline
		FCN-$8$s~\cite{long2015fully} & VGG-$16$ & - & $35.1$ \\
		BoxSup~\cite{dai2015boxsup} & - & - & $40.5$ \\
		DeepLab-v$2$~\cite{chen2018deeplab} & Dilated-ResNet-$101$ & -& $45.7$ \\
		RefineNet~\cite{lin2017refinenet} & ResNet-$152$ & - & $47.3$ \\
		PSPNet~\cite{zhao2017pyramid} & Dilated-ResNet-$101$ & $47.8$ & - \\
		Ding et al.~\cite{ding2018context} & ResNet-$101$ & $51.6$ & - \\
		EncNet~\cite{zhang2018context} & Dilated-ResNet-$101$ & $52.6$ & - \\
		HRNetV$2$\cite{sun2019high} & HRNetV$2$-W$48$ &  $54.0$ & $48.3$ \\	
		\hline
		Our approach & GSTO-HRNet-W$48$ &  $\mathbf{54.3}$ & $\mathbf{48.5}$ \\	
		\hline
	\end{tabular}
	\caption{Semantic segmentation results on PASCAL-context. The methods are
	evaluated on $59$ classes and $60$ classes.}
	\label{tab:pasctx_sota}
	\end{table}

\subsection{Dataset}
\subsubsection{Semantic Segmentation}

\quad\textit{Cityscapes}
\cite{cordts2016cityscapes}. Cityscapes is a large-scale dataset focusing on semantic understanding of urban street scenes, containing $5,000$ pixel-level annotated scene images divided into $2,975/500/1,525$ images for training, validation and testing, respectively. For pixel-level labeling, there are $30$ classes annotated and $19$ of them used for evaluation.  

\textit{LIP}
\cite{gong2017look}. LIP is an elaborately annotated human parsing dataset, which contains $50,462$ images annotated with $20$ categories ($19$ for human parts and $1$ for background). There are $30,462$ images for training and $10,000$ for validation. 

\textit{Pascal Context}
\cite{mottaghi2014role}. Pascal Context is a challenging scene parsing dataset, including $4,998$ images for training and $5,105$ for testing. There are $60$ classes, $59$ for semantic category and $1$ for background. We have tested our model on both conditions whether or not to ignore the background (denoted as ``$59$ classes" and ``$60$ classes", respectively). 

\subsubsection{Pose Estimation}

\quad\textit{COCO}
\cite{lin2014microsoft}. We use the COCO \texttt{train$2017$} for training, which contains $57K$ images and $150K$ person instances, annotated with 17 keypoints. Then we evaluate our model on COCO \texttt{val}$2017$ and \texttt{test-dev}$2017$.

\subsection{Evaluation metric}
\subsubsection{Semantic Segmentation}
We report the result for semantic segmentation mainly on the IoU-based metrics. IoU (intersection-over-union) is calculated by $TP/(TP+FP+FN)$, where $TP$, $FP$ and $FN$ are the numbers of true positive, false positive and false negative pixels, respectively. For standard evaluation, mIoU (mean of IoU among based on classes) is exploited. 

\subsubsection{Pose Estimation}
OKS-based mAP (AP for short) is used for human pose estimation. OKS (Object Keypoint Similarity) is calculated through an unnormalized Gaussian distribution and outputs a value between $0$ and $1$, representing the similarity between the prediction and ground truth. Following previous arts, we report the average precision on several cases including large objects ($\operatorname{AP}^{L}$), medium objects ($\operatorname{AP}^{M}$), $\operatorname{OKS}=0.5$ ($\operatorname{AP}^{50}$), $\operatorname{OKS}=0.75$ ($\operatorname{AP}^{75}$), mean of $\operatorname{OKS}=0.50,0.55,...,0.95$, and the mean of average recall score on $\operatorname{OKS}=0.50,0.55,...,0.95$ ($\operatorname{AR}$).

\subsection{Implementation Details}
\subsubsection{Semantic Segmentation.}
We follow the training protocol in \cite{sun2019high}. For data augmentation, random cropping ($512\times 1024$ for Cityscapes, $473\times 473$ for LIP and $480\times 480$ for Pascal Context), random scaling in the range of $[0.5,2]$ and random horizontal flipping are exploited. We use SGD with momentum of $0.9$ and poly learning rate policy with the power of $0.9$ for all the datasets. The base learning rate is set as $0.01$ for Cityscapes, $0.007$ for LIP and $0.004$ for Pascal Context. Weight decay is set as $0.0005$ for Cityscapes and LIP, and $0.0001$ for Pascal Context. Besides, we train our models for $600$ epochs on Cityscapes with batch size of $12$, $150$ epochs on LIP with batch size of $40$ and $280$ epochs on Pascal Context with batch size of $16$.
\subsubsection{Pose Estimation.} Following \cite{sun2019deep}, we resize the detected box to fixed size: $384\times 288$. Random rotation in the range of $[-45^\circ , 45^\circ]$, random scale in the range of $[0.65, 1.35]$ and flipping are exploited as augmentation. We use the Adam optimizer and the base learning rate is set as $0.001$, reduced to $0.0001$ and $0.00001$ at the $170$th and $200$th epochs, respectively. The batch size is set $128$ for GSTO-HRNet-W$32$ and $80$ for GSTO-HRNet-W$48$. 

\subsection{Ablation Study}
\label{sec:4.3}
We evaluate the proposed GSTO on the strong multi-scale backbone HRNet and conduct experiments to compare different inserting strategies and locations. Unless explicitly noted, the baseline model in this section is HRNetV$2$-W$18$-Small-v$2$, the results are reported on the Cityscapes validation dataset and the GFLOPs are calculated on the input size of $1024 \times 2048$. 

\subsubsection{GSTO-based modules}
Firstly, we evaluate two GSTO-based modules: Gated Fusion Module (GFM) and Gated Transition Module (GTM). As described in Section~\ref{sec3.2}, GFM is exploited after each block to adaptively fuses the multi-scale features from each branch, and GTM is exploited after each stage to adaptively generate a lower-resolution branch. As shown in Table~\ref{tab:ablation_hrnet1}, performing GFM and unsupervised GTM improves the baseline by $0.7\%$ and $0.5\%$ respectively, and by $0.9\%$ when both of them are applied. When we combine GTM with the auxiliary supervision as proposed in Section~\ref{sec3.2}, the result booms to $77.0\%$. When cooperated with GFM, it achieves $1.3\%$ enhancement compared with the baseline ($75.9\% \rightarrow 77.2\%$). \textit{Specially, we note that simply utilizing extra supervision can only gain $0.4\%$ growth, indicating that the proposed supervised GSTO makes good use of auxiliary supervision. }

\subsubsection{Number and location of supervised GTM}
we further explore the number and location to apply the supervised GTM. We set the baseline as HRNetV$2$ with GFM only. Table~\ref{tab:ablation_hrnet2} shows that auxiliary supervision is useful, and when the number of supervised GTM increases from $1$ to $2$, the performance improves from $76.7\%$ to $77.3\%$. But exploiting to all three stages leads to $0.1\%$ drop, indicating that shallow layers lacking semantic information can be even harmful for conducting the generation of the spatial gate. Therefore, exploiting supervised GTM only in stage$2$ and stage$3$ is the best choice with limited extra parameters and FLOPs, and we name this model as GSTO-HRNet.

\subsubsection{Compare with other gating mechanisms}
In this part, we compare the proposed GSTO with previous attention-based modules, demonstrating that GSTO is a more effective strategy to apply gating/attention mechanism into pixel-labeling networks. 
We conduct experiments to compare GSTO (spatial attention combined with scale-transfer operation) with other attention forms (channel attention) and other locations to plug into the modules (after each block or after each layer). The results (as shown in Table\ref{table:coco}) demonstrate that spatial attention is much more effective than channel-wise for pixel-labeling tasks, and \textit{combining with scale-transfer operations for multi-scale feature extraction is a better way to bring the superiority of gating/attention mechanism than directly utilizing it in the end.}

\begin{figure*}[t]
  \includegraphics[width=\textwidth]{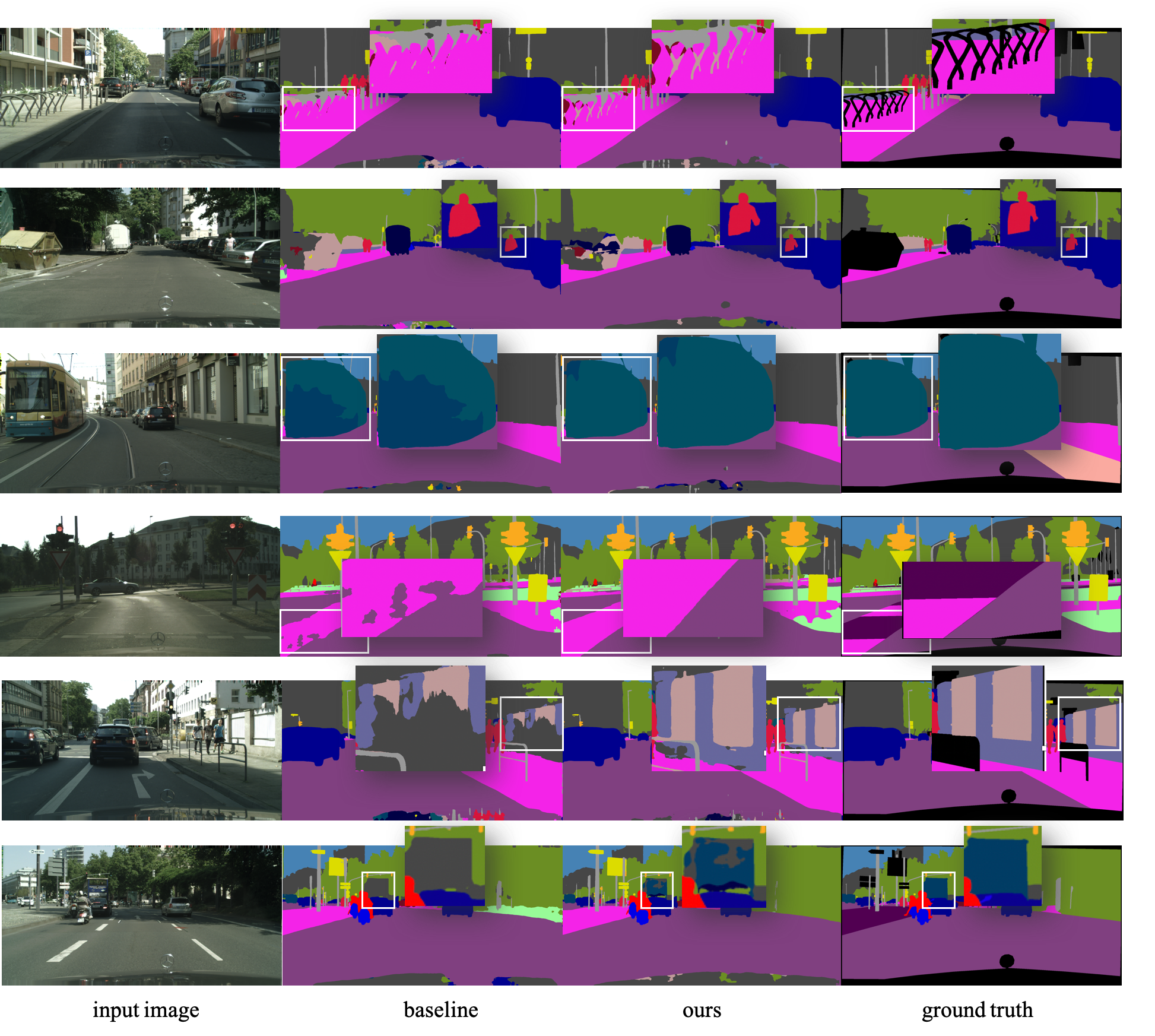}
  \caption{Example segmentation results of our method GSTO-HRNet-W$48$ and baseline HRNetV$2$-W$48$.}
  \label{fig:vis}
\end{figure*}

\subsection{Results on baseline}
As demonstrated, the proposed GSTO can be easily inserted into any multi-scale feature extraction modules or backbones. We evaluate our method on strong multi-scale backbone HRNet\cite{sun2019high}\cite{sun2019deep} for both semantic segmentation and pose estimation, and on typical multi-scale modules PPM\cite{zhao2017pyramid} and ASPP\cite{chen2018deeplab} for semantic segmentation.

\subsubsection{Results on GSTO-HRNet}
We compare the proposed GSTO-HRNet with the baseline model HRNet in the aspects of parameters, computational costs and accuracy on Cityscapes for semantic segmentation and COCO for human pose estimation.

For Semantic segmentation, as presented in Table~\ref{tab:ablation_hrnet3}, with extra parameters less than $1\%$ ($0.67\%$ for HRNetV$2$-W$18$ and $0.23\%$ for HRNetV$2$-W$48$) and GFLOPs less than $5\%$ ($3.9\%$ for HRNetV$2$-W$18$ and $2.6\%$ for HRNetV$2$-W$48$), our GSTO-HRNet booms the mIoU by a large margin ($75.9\% \rightarrow 77.3\%$ and $80.2\% \rightarrow 82.1\%$, respectively) on the Cityscapes validation set. While in Table~\ref{tab:ablation_hrnet4}, exhaustive experiments on Cityscapes test set are conducted to compare the performance of HRNet and our approach, and the results demonstrate the superiority of our GSTO-HRNet.

For human pose estimation results in Table~\ref{table:coco}, our GSTO-HRNet-W$32$ significantly enhances the baseline from $75.8\%$ AP to $76.5\%$ AP on COCO validation set, which even outperforms the HRNet-W$48$ by $0.2$ points while saving $53\%$ parameters and $45\%$ GFLOPs. The proposed GSTO-HRNet-W$48$ further improves the AP to $76.7\%$.

\subsubsection{Results on Multi-scale Modules with GSTO}
Besides multi-scale backbones, the proposed Gated Scale-Transfer Operation can also improve general multi-scale modules by simply replacing conventional scale-transfer operation with GSTO. We verify this on the Cityscapes validation dataset for semantic segmentation in Table~\ref{tab:ablation_ppm}. The backbone is set as ResNet-$50$. It shows that supervised GSTO advances PPM by $1.3$ points ($76.5\%\rightarrow 77.8\%$) and advances ASPP by $2.0$ points ($74.9\% \rightarrow 76.9\%$). 

\subsection{Comparison with state-of-the-art}
The proposed GSTO-HRNet achieves new state-of-the-art results on multiple benchmarks for pixel labeling tasks. 

For semantic segmentation, GSTO-HRNet achieves $81.8\%$ mIoU on Cityscapes test set using \textbf{only fine-labeled} train set and $82.4\%$ using \textbf{only fine-labeled} train-val set (Table~\ref{tab:cityscaperesults}), which is the highest performance without using extra data like \textit{Mapillary} and \textit{COCO}. Besides, on the human parsing dataset LIP, GSTO-HRNet improves the  state-of-the-art method by $1.5\%$ on mIoU (Table~\ref{tab:lipresults}), and on the complex scene parsing dataset Pascal Context, it again achieves best performance on both $59$ classes and $60$ classes (Table~\ref{tab:pasctx_sota}). 

For pose estimation, we use the same person detector and tracking strategy as \cite{sun2019deep}. As shown in Table~\ref{table:coco}, performed on challenging COCO dataset without using any extra training data, our approach reaches state-of-the-art AP $76.7\%$ on the validation set and $75.8\%$ on the test set. Impressively, GSTO-HRNet achieves state-of-the-art result on the validation set ($76.5\%$), outperforming the original HRNet-W$48$ with less parameters and smaller computational cost.

\section{Further Visualization}
Figure~\ref{fig:introduction} has illustrated the visual comparison of the multi-scale features extracted by the encoder of HRNetV$2$-W$48$ (as the baseline) and the proposed GSTO-HRNet-W$48$. In this section, we further provide the qualitative comparisons of segmentation results in Figure~\ref{fig:vis}. 

It indicates that our method captures more accurate boundary details than baseline model, like the road rail in the first instance and the leg of the person in the second row. 
Besides, the proposed GSTO keeps a consistency on large objects, like the "sidewalk" in the third instance, and the huge "train" in the fouth instance. 
Morever, our method clearly obtains better semantic comprehension, since in the fifth example, the baseline method can hardly distinct the "fence", "wall" and "building", while our method achieves much better result. And in the last row, the occluded "bus" can be easily confused with "truck", for which our model also performs better. 

\section{Conclusion}
In this paper, we have proposed two forms of Gated Scale-Transfer Operations (GSTOs) for extracting more discriminative and scale-aware multi-scale features in pixel labeling. Experiments show that GSTOs significantly boost the multi-scale backbone HRNet and multi-scale modules like PPM and ASPP, with negligible extra parameters and computational cost. Moreover, the GSTO-based architecture GSTO-HRNet achieves new state-of-the-art results on Cityscapes, LIP and Pascal Context datasets for semantic segmentation, and COCO for pose estimation.

{\small
\bibliographystyle{ieee}
\bibliography{gsto}
}

\newpage

\end{document}